%% file: main.tex
\documentclass[letterpaper]{article} % DO NOT CHANGE THIS
\usepackage{aaai25}  % DO NOT CHANGE THIS
\usepackage{times}  % DO NOT CHANGE THIS
\usepackage{helvet}  % DO NOT CHANGE THIS
\usepackage{courier}  % DO NOT CHANGE THIS
\usepackage[hyphens]{url}  % DO NOT CHANGE THIS
\usepackage{graphicx} % DO NOT CHANGE THIS
\usepackage{natbib}  % DO NOT CHANGE THIS AND DO NOT ADD ANY OPTIONS TO IT
\usepackage{caption} % DO NOT CHANGE THIS AND DO NOT ADD ANY OPTIONS TO IT
\usepackage{algorithm}
\usepackage{algorithmic}
\usepackage{amsmath}
\usepackage{graphicx}
\usepackage{subcaption}
\usepackage{pifont}   % 提供 \ding 命令，用于插入特殊符号（如勾号）
\usepackage{makecell} % 提供 \makecell 命令，用于在单元格中换行和格式化
\usepackage{newfloat}
\usepackage{listings}
\DeclareCaptionStyle{ruled}{labelfont=normalfont,labelsep=colon,strut=off} % DO NOT CHANGE THIS
\floatstyle{ruled}
\newfloat{listing}{tb}{lst}{}
\floatname{listing}{Listing}
\title{JAQ: Joint Efficient Architecture Design and Low-Bit Quantization with \\ Hardware-Software Co-Exploration}

\author{
    Mingzi Wang\textsuperscript{\normalfont 1},
    Yuan Meng\textsuperscript{\normalfont 2*},
    Chen Tang\textsuperscript{\normalfont 1,2},
    Weixiang Zhang\textsuperscript{\normalfont 1},
    Yijian Qin\textsuperscript{\normalfont 2},
    Yang Yao\textsuperscript{\normalfont 2},
    Yingxin Li\textsuperscript{\normalfont 1},
    Tongtong Feng\textsuperscript{\normalfont 2},
    Xin Wang\textsuperscript{\normalfont 2},
    Xun Guan\textsuperscript{\normalfont 1*},
    Zhi Wang\textsuperscript{\normalfont 1*},
    Wenwu Zhu\textsuperscript{\normalfont 2*}
}
\affiliations{
    \textsuperscript{\rm 1}Tsinghua Shenzhen International Graduate School, Tsinghua University, Shenzhen, China\\
    \textsuperscript{\rm 2}Department of Computer Science and Technology \& BNRist, Tsinghua University, Beijing, China \\
wmz22@mails.tsinghua.edu.cn, yuanmeng@mail.tsinghua.edu.cn, \{xun.guan,wangzhi\}@sz.tsinghua.edu.cn,
wwzhu@tsinghua.edu.cn
}

\usepackage{bibentry}
\begin{document}

\maketitle

\begin{abstract}
The co-design of \textit{neural network architectures, quantization precisions, and hardware accelerators} offers a promising approach to achieving an optimal balance between performance and efficiency, 
particularly for model deployment on resource-constrained edge devices.
In this work, we propose the \textbf{JAQ} Framework, which jointly optimizes the three critical dimensions. However, effectively automating the design process across the vast search space of those three dimensions poses significant challenges, especially when pursuing extremely low-bit quantization. Specifical, the primary challenges include:  (1) \textit{Memory overhead in software-side}: Low-precision quantization-aware training can lead to significant memory usage due to storing large intermediate features and latent weights for back-propagation, potentially causing memory exhaustion. (2) \textit{Search time-consuming in hardware-side}: The discrete nature of hardware parameters and the complex interplay between compiler optimizations and individual operators make the accelerator search time-consuming. To address these issues, 
\textbf{JAQ} mitigates the memory overhead through a channel-wise sparse quantization (\textbf{CSQ}) scheme, selectively applying quantization to the most sensitive components of the model during optimization. Additionally, \textbf{JAQ} designs \textbf{BatchTile}, which employs a hardware generation network to encode all possible tiling modes, thereby speeding up the search for the optimal compiler mapping strategy. Extensive experiments demonstrate the effectiveness of \textbf{JAQ}, achieving approximately 7\% higher Top-1 accuracy on ImageNet compared to previous methods and reducing the hardware search time per iteration to 0.15 seconds. 
Code is available at {https://github.com/wmz-opensource/JAQ/}. 
\end{abstract} 

\section{Introduction}
\label{sec:intro}
\input{intro}

\section{Related Work}
\label{sec:related}
\input{related}

\section{JAQ Framework}
\label{sec:methods}

\input{methods}

\section{Experiments}
\label{sec:experiments}
\input{experiments}

\section{Conclusion}
\label{sec:conclusion}
\input{conclusion}

\section{Acknowledgments}
This work was supported by National Key Research and Development Project of China~(Grant No. 2023YFF0905502), National Natural Science Foundation of China~(Grant No. 62472249, No. 62402264), Shenzhen Science and Technology Program~(Grant No. JCYJ20220818101014030), National Key R\&D Program of China~(2021YFA1001000), National Natural Science Foundation of China~(U23A20282), Shenzhen Science, Technology and Innovation Commission (KJZD20231023094659002, JCYJ20220530142809022, WDZC20220811170401001). 
We thank anonymous reviewers for their valuable advice.

\bibliography{paper}

\clearpage
\section{Appendix}

\subsection{A\hspace{1em}Memory Cost Measurement}
All experiments are conducted on a single NVIDIA GeForce RTX 4090 GPU to prevent inaccuracies in memory measurements caused by multiple copies of the model in multi-GPU parallel setups. We find that as the batch size increases linearly, the GPU memory utilization also increases linearly. Therefore, by measuring memory cost with a small batch size, we can estimate the memory utilization for larger batch sizes. We utilize FBNet~\cite{wu2019fbnet} as the supernet, and add quantization process(Eq.~\ref{eq:quantization}) in each operator.

\subsection{B\hspace{1em}Parameter coupling or misguided search}
\begin{figure}[h]
    \centering
    % First image
    \begin{subfigure}[b]{0.23\textwidth}
        \centering
        \includegraphics[width=1.5in, height=1.3in]{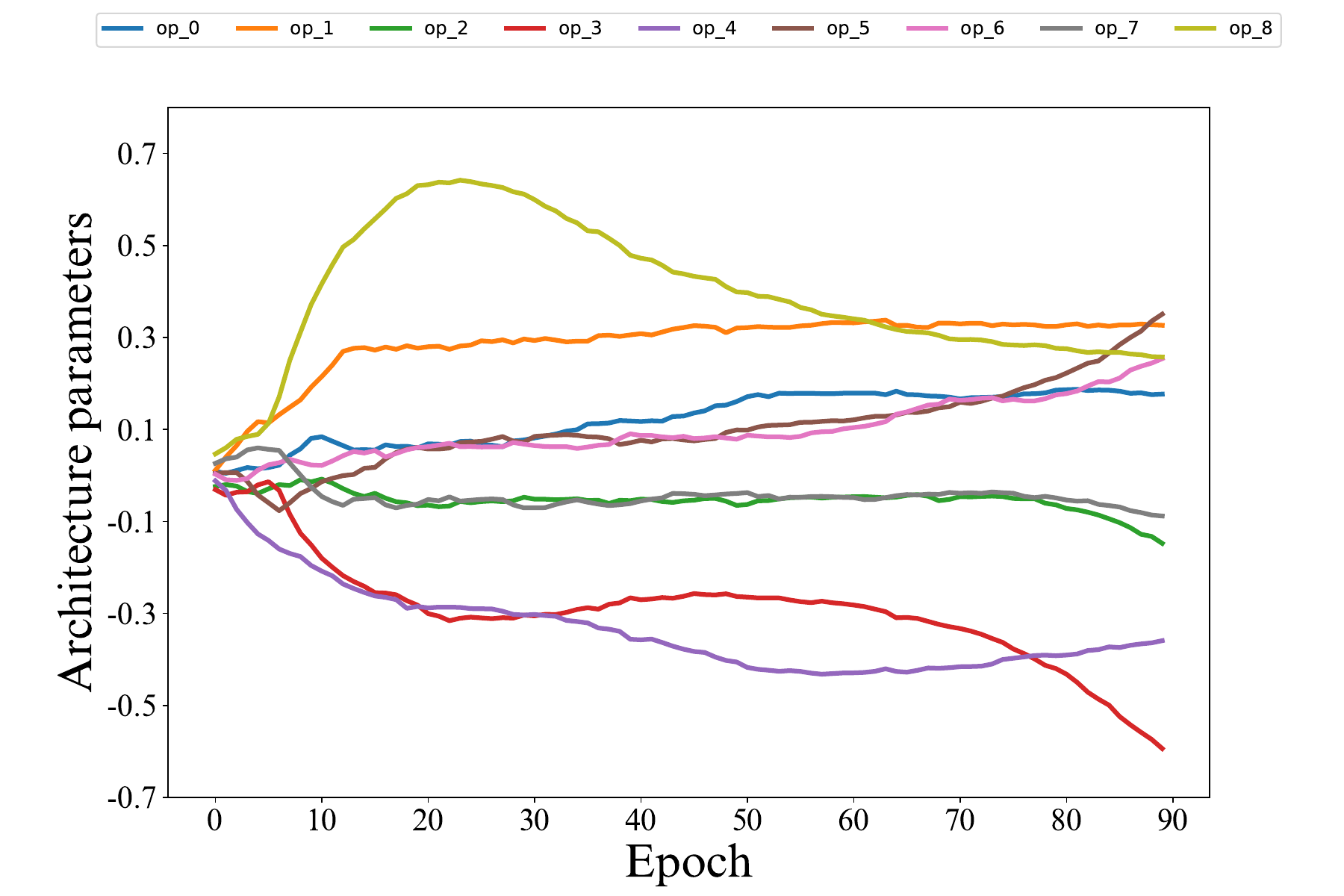} % Adjust path as needed
        \captionsetup{font=tiny}
        \caption{parameter coupling}
        \label{fig:parameter coupling}
    \end{subfigure}
    \hfill
    \begin{subfigure}[b]{0.23\textwidth}
        \centering
        \includegraphics[width=1.5in, height=1.3in]{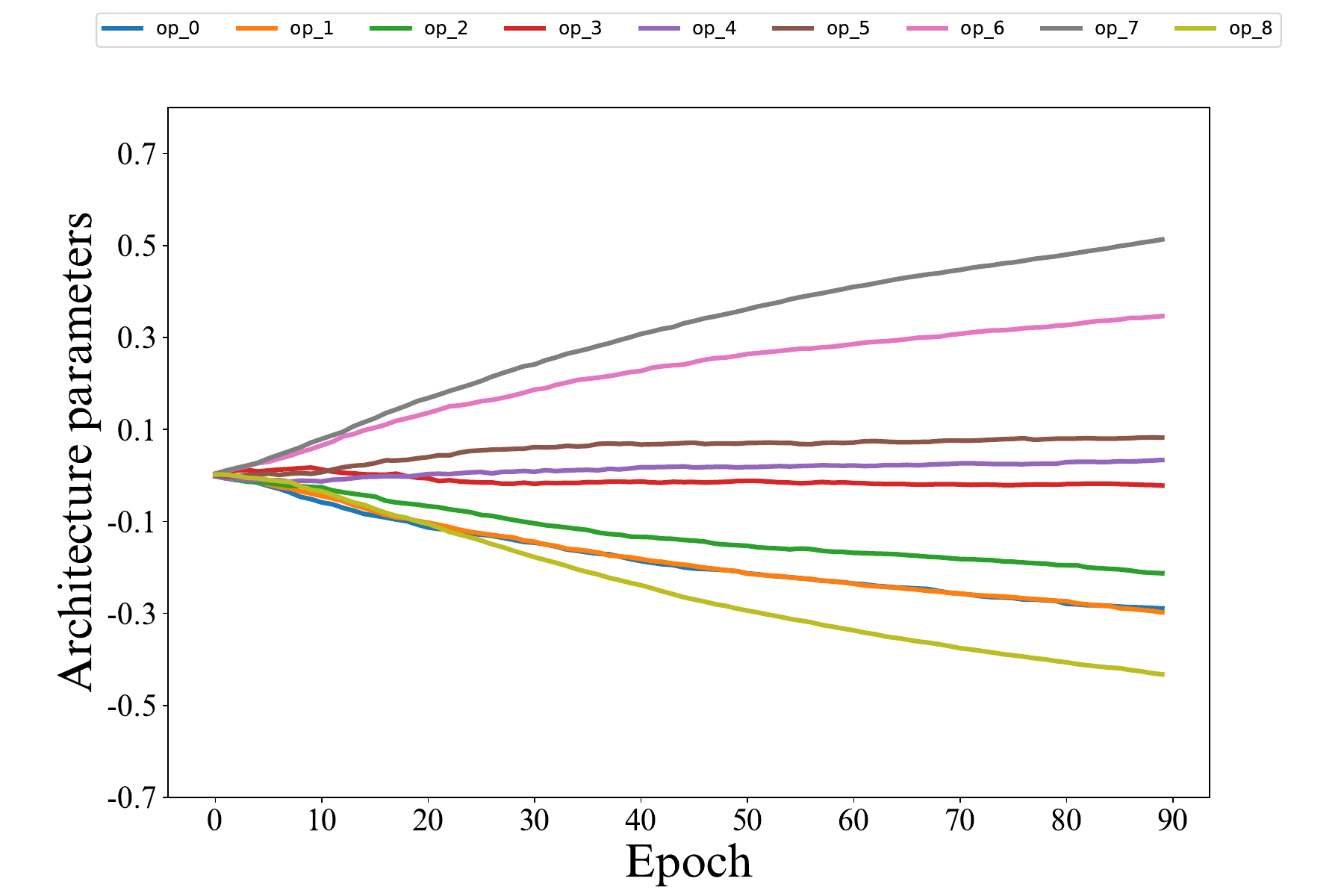} % Adjust path as needed
        \captionsetup{font=tiny}
        \caption{w/o parameter coupling}
        \label{fig:no parameter coupling}
    \end{subfigure}
    \hfill
    % Second image
    \begin{subfigure}[b]{0.23\textwidth}
        \centering
        \includegraphics[width=1.5in, height=1.3in]{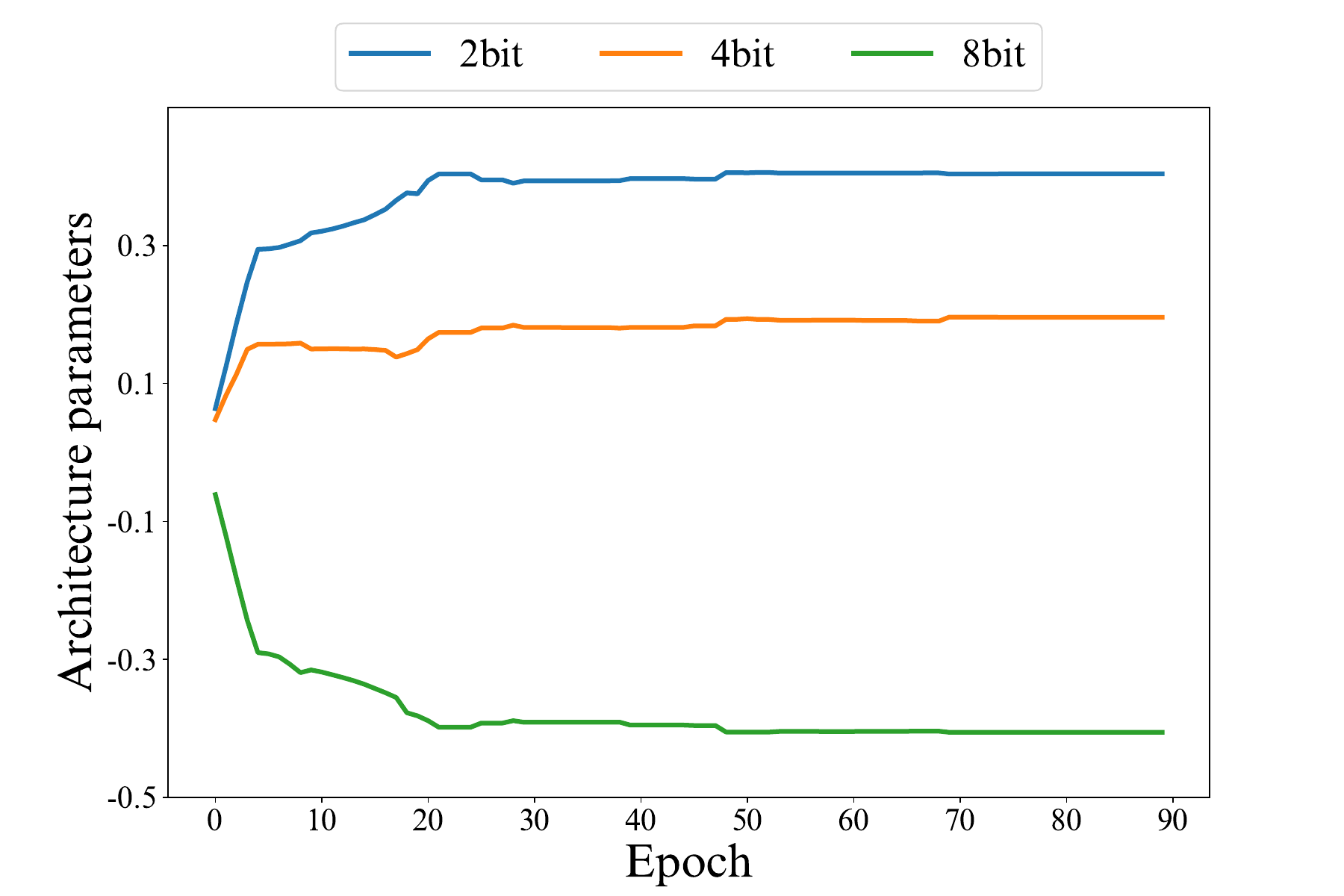} % Adjust path as needed
        \captionsetup{font=tiny}
        \caption{misguided search}
        \label{fig:misguided search}
    \end{subfigure}
    \hfill
    % First image
    % First image
    \begin{subfigure}[b]{0.23\textwidth}
        \centering
        \includegraphics[width=1.5in, height=1.3in]{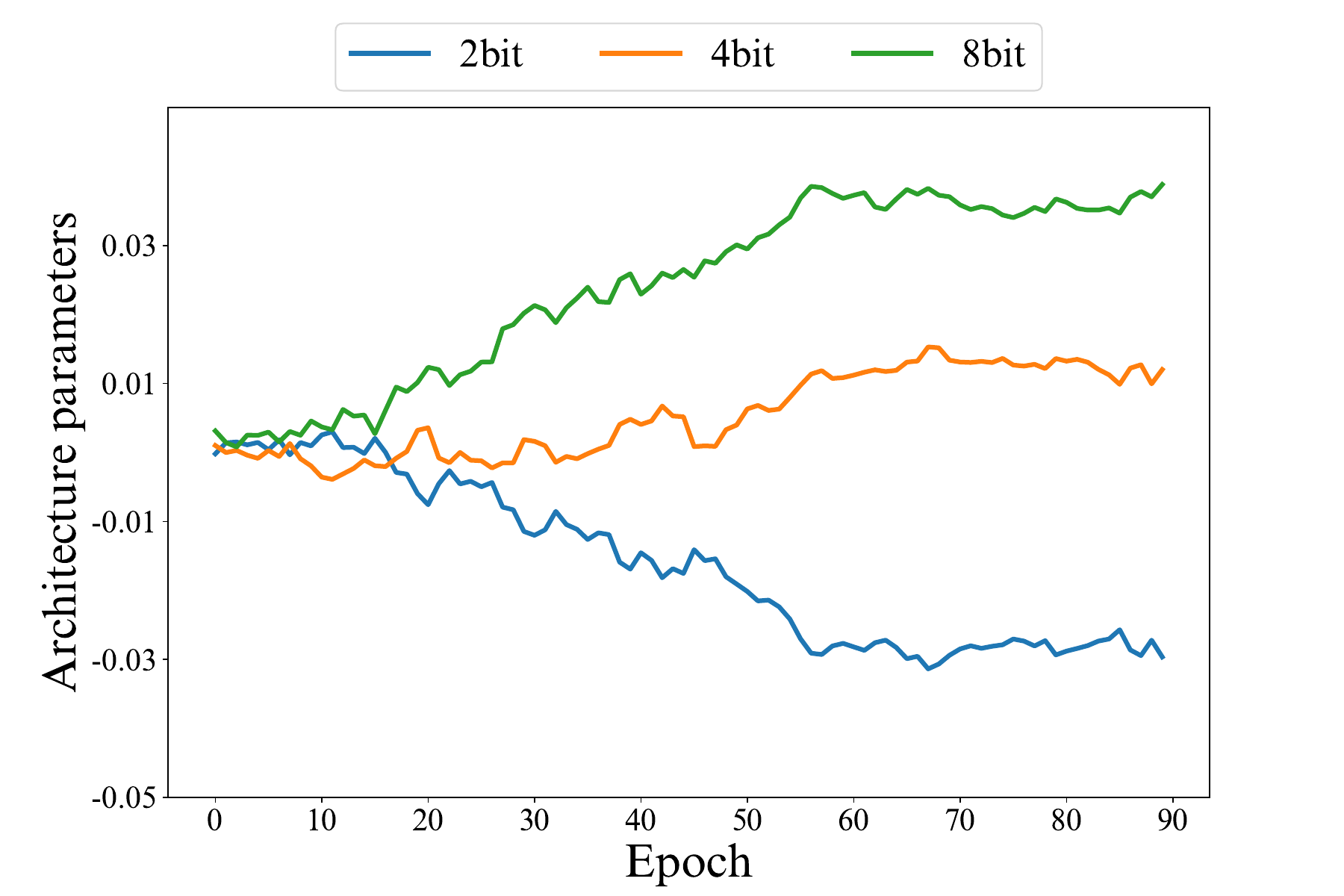} % Adjust path as needed
        \captionsetup{font=tiny}
        \caption{w/o misguided search}
        \label{fig:no misguided search}
    \end{subfigure}

    \caption{(a) and (c) demonstrate the problems of parameter coupling and misguided search in the previous work~\cite{fu2021auto} algorithm under unconstrained condition. (b) and (d) illustrate that our channel-wise sparse quantization algorithm does not exist parameter coupling or misguided search issues under unconstrained condition. }
    % \yym{the size of legend is very strange}}
    \label{fig:bias_search}
\end{figure}

\subsection{C\hspace{1em}Accelerator Search Time Comparison}

\begin{table}[H]
\centering

\label{search time}
\resizebox{0.3\textwidth}{!}{ % 控制表格的大小
\renewcommand{\arraystretch}{0.4} % 控制行距
\LARGE
\begin{tabular}{cc}
\\ \cline{1-2} \\
\multicolumn{1}{c}{\bf \makecell{Method}}
&\multicolumn{1}{c}{\bf \makecell{Search Time}}
\\ \cline{1-2} \\
\makecell{Auto-nba~\cite{fu2021auto}}
&\makecell{30}
\\ \cline{1-2} \\
\makecell{\textbf{Ours}}
&\makecell{0.15}
\\ \cline{1-2} \\
\end{tabular}
}
\vspace{0.1cm}
\caption{Comparison of Accelerator search time(s) between JAQ and previous work.}
\label{tab:accelerator search time}
\end{table}

\end{document}

%% file: intro.tex
Given the significant computational demands of Deep Neural Networks (DNNs), deploying them in resource-limited environments, such as the Internet of Things (IoT), remains a challenge. For example, even highly optimized Convolutional Neural Networks (CNNs) have recently struggled to perform efficiently on resource-constrained hardware devices \cite{li2023pareto,rashid2024tinym2net}. 
To speed up inference on real-world hardware while maintaining performance, hardware-aware techniques (e.g., quantization, and hardware-aware neural architecture search) have emerged to improve the model efficiency on the model-side. 
For example, HAQ~\cite{wang2019haq}, OFA~\cite{cai2019once}, and ElasticViT~\cite{tang2023elasticvit} optimize the model for a fixed target device. 
On the other hand, accelerator-side methods design specialized accelerators~\cite{chen2014dadiannao,liu2015pudiannao,parashar2017scnn} to facilitate the deployment of DNNs, have received more attention recently. 

However, the separated design on either the model-side or accelerator-side falls into sub-optimal \cite{fu2021auto,hong2022enabling} as (1) the model-size optimization will be up against efficiency loss when the hardware does not support certain operators and (2) the optimal accelerator design varies very different for various model structures and the corresponding quantized precision \cite{wang2019haq}. 
This trend suggests the limitations of and the need for co-design of both neural networks, quantized bit-widths, and hardware accelerators.

The first principle of co-design involves efficiently navigating the vast design space. To achieve this, differentiable methods have been developed to facilitate end-to-end co-exploration. 
Notably, AutoNBA~\cite{fu2021auto} utilizes learnable weights for determining the expected precision and architectural operator, along with designing a new objective for optimizing hardware components. 
DANCE~\cite{choi2021dance} further introduces an MLP-based accelerator search strategy into the differentiable search framework. 
However, these methods support only high bit-width quantization (i.e., $\geq$ 4 bits), resulting in minimal performance degradation due to the significant redundancies that remain for compression \cite{esser2019learned}. 
In contrast, we have observed that low-precision disrupts the optimization process, leading to a misguided search, as we will discuss later.

Therefore, we propose JAQ framework, which addresses challenges and achieves efficient joint exploration.
For the first challenge, we propose channel-wise sparse quantization method. It selects a small subset of the most crucial activations channels for quantization, leaving other channels unquantized during the search process, effectively alleviating the issue of memory explosion.
For the second challenge, we propose BatchTile approach that encodes all tile sizes within the search space as different batches, enabling us to determine the optimal tiling strategies simultaneously, which significantly reduces the time overhead.

\begin{table}[t]
\centering

\label{search time}
\resizebox{0.45\textwidth}{!}{ % 控制表格的大小
\renewcommand{\arraystretch}{0.7} % 控制行距
\LARGE
\begin{tabular}{cccc}
% \toprule[1.2pt]
\\ \cline{1-4} \\
\multicolumn{1}{c}{\bf \makecell{Method}} 
&\multicolumn{1}{c}{\bf \makecell{Model \\ Achitecture}}
&\multicolumn{1}{c}{\bf \makecell{Low-Precision \\ Quantization}}  
&\multicolumn{1}{c}{\bf \makecell{Accelerator \\ Architecture }}
\\  \cline{1-4} \\
\makecell{NAAS~\cite{lin2021naas}}
&{\makecell{\ding{53}}}
&{\makecell{\ding{53}}}
&{\makecell{\ding{51}}}
\\ \cline{1-4} \\
\makecell{DANCE~\cite{choi2021dance}}
&\makecell{\ding{51}}
&\makecell{\ding{53}}
&\makecell{\ding{51}}
\\ \cline{1-4} \\
\makecell{Auto-nba~\cite{fu2021auto}}
&\makecell{\ding{51}}
&\makecell{\ding{53}}
&\makecell{\ding{51}}
\\ \cline{1-4} \\
\makecell{\textbf{Ours}}
&\makecell{\ding{51}}
&\makecell{\ding{51}}
&\makecell{\ding{51}}
\\ \cline{1-4} \\
% \\ \bottomrule[1.2pt]
\end{tabular}
}
\vspace{0.1cm}
\caption{Comparison with other works on the search space dimension.}
\label{tab:intro}
\end{table}

To summarize, the contributions of the paper are:
\begin{itemize}
    \item We propose the JAQ framework, which enables efficient and effective co-exploration within the extensive optimization space. To the best of our knowledge, we are the first to explore the joint search among network architecture, ultra-low mixed-precision bit-width allocation, and accelerator architecture, as shown in Tab.~\ref{tab:intro}. 
    \item To tackle the challenge of memory explosion, we propose the channel-wise sparse quantization approach, achieving around 5$\times$ reduction in memory cost compared to non-optimized scenarios. 
    \item We propose a hardware generation network to optimize accelerator design and BatchTile method to integrate the compiler mapping search efficiently, which reduces the search time per iteration to 0.15 seconds.
    \item Extensive experimental evaluations demonstrate that our framework surpasses the state-of-the-art. Our work opens up new possibilities for agile software-hardware co-design.
\end{itemize}

%% file: related.tex
\subsection{Quantization and Neural Architecture Search}

As a hardware-friendly lightweight technique, quantization has broad prospects for application.
Mixed-Precision Quantization (MPQ)~\cite{dong2019hawq,wang2019haq,tang2022mixed,kim2024metamix,huang2022sdq} allocates different bitwidths to the activations and weights of each layer, showing better accuracy-efficiency trade-off compared to fixed-precision quantization~\cite{choi2018pact,esser2019learned,markov2023quantized,xu2023q,nagel2022overcoming}. Recently, hardware increasingly supports mixed-precision~\cite{sharma2018bit,umuroglu2018bismo}, which further pushes the research in MPQ. HAQ~\cite{wang2019haq} leverages Reinforcement learning (RL) to allocate bitwidth to each layer. HAWQ~\cite{dong2019hawq} uses the information derived from the Hessian matrix to determine quantization sensitivity and guide the allocation of bitwidths for network parameters.

Neural Architecture Search (NAS) enables the automated design of high-performance DNN network structures, saving time and effort of the manual design. 
To reduce search cost, differentiable NAS~\cite{liu2018darts, qin2021graph} methods have merged, which integrate all candidate operators into an end-to-end trained supernet, and finally select the optimal subnet. Some studies have incorporated hardware performance metrics into the NAS via lookup tables~\cite{zhang2020fast,li2021hw}, aiming to enhance the model's efficiency on actual hardware. However, all these works concentrate exclusively on algorithmic optimization without exploring hardware architecture, which may not yield optimal inference efficiency.

\subsection{DNN Accelerators}

To improve the performance of modern deep neural network computations, fixed-bitwidth DNN accelerators have emerged, featuring specialized components like MAC arrays, on-chip buffers, and network-on-chip 
architectures~\cite{chen2016eyeriss,jouppi2017datacenter,du2015shidiannao}. 
Recently, the concept of MPQ has paved the way for the development of bit-flexible accelerators~\cite{sharma2018bit,umuroglu2018bismo} that allow for varying bitwidths across individual layers. However, designing AI accelerators remains a complex and time-consuming task that demands significant hardware expertise. 
However, designing AI accelerators is complex and requires significant expertise. AI-driven methods, such as NAAS~\cite{lin2021naas} and GPT4AIGChip~\cite{fu2023gpt4aigchip}, streamline the process by autonomously evaluating design configurations. These approaches focus primarily on hardware architecture and often yield sub-optimal results compared to co-design methodologies that integrate both network and hardware exploration~\cite{you2023vitcod,lou2023naf,stevens2021softermax,reggiani2023mix}.

\subsection{Hardware-software Co-design}

\begin{figure*}[t]
    \centering
    \includegraphics[width=0.95\linewidth]{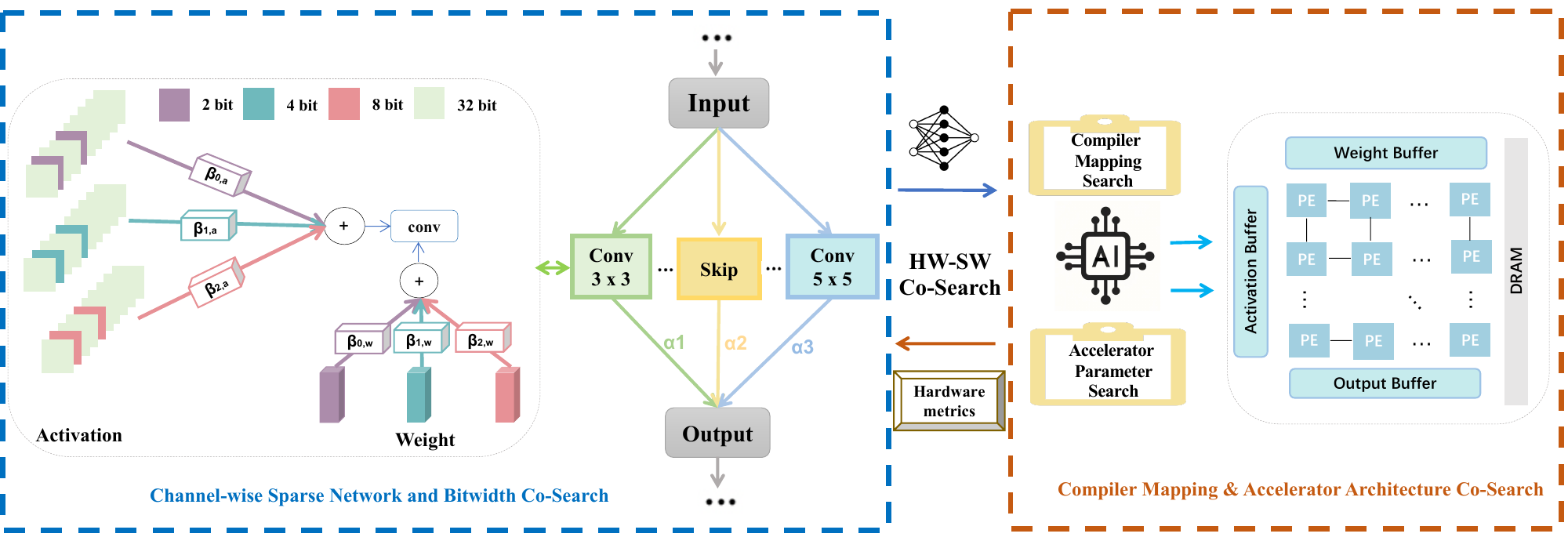}
    \caption{JAQ framework. The left part represents the optimization of network structure and bitwidths allocation, addressing the memory cost bottleneck through channel-wise sparse quantization. The right part depicts accelerator architecture search, including hardware parameters and compiler mapping strategy. Hardware metrics indicate accelerator performance (Energy, Latency and Area).}
    \label{fig:main}
\end{figure*}
Some studies employ hardware-software co-design methods using reinforcement learning or evolutionary algorithm~\cite{jiang2020hardware,abdelfattah2020best}, which require expensive training time and also suffer from limited search spaces. To address this issue, differentiable methods have been employed for co-exploration. EDD~\cite{li2020edd} is an FPGA-based differentiable network design framework. However, it does not encompass the search for hardware parameters, such as the number of BRAMs or DSPs. While Dance~\cite{choi2021dance} builds a pipeline to explore ASIC-based accelerator and network structure, it has a limitation that it does not take quantization into consideration.
Auto-nba~\cite{fu2021auto} is not suitable for low-bit quantization.

JAQ targets efficient joint search of network, low-bit mixed-precision bitwidths and accelerator architecture.

%% file: methods.tex
\subsection{Preliminary}
\label{subsec:preliminary}
\paragraph{Differentiable Neural Architecture Search.}
Differentiable neural architecture search (DNAS) \cite{liu2018darts,wu2019fbnet} transforms the entire search space into a supernet and each path in the supernet is equipped with an architecture parameter, which represents the probability of selecting this path.
The incorporation of the Gumbel-Softmax\cite{jang2016categorical} function plays a pivotal role enabling these architecture parameters trainable through gradient-based optimization. After the training of the supernet, the optimal subnet is formed by the path with the highest architecture parameter in each layer. The function of Gumbel-Softmax is:

\begin{gather}
\beta_t = \frac{\exp\left(\frac{\beta_t + \epsilon_t}{\tau}\right)}{\sum_{i=1}^N \exp\left(\frac{\beta_i + \epsilon_i}{\tau}\right)}, \quad \epsilon \sim U(0,1), \label{eq:gumbel}
\end{gather}
where $\beta$ represents the original parameter distribution, while $\epsilon$ is a number sampled from a uniform distribution ranging between 0 and 1. Additionally, the smoothness of the distribution can be regulated using the temperature coefficient $\tau$.

\paragraph{Quantization.}
The quantization function $Q(\cdot)$, defined as:
\vspace{0.05cm}
\begin{gather}
Q(V) = \text{round} \left( \text{clip} \left( \frac{V}{s}, \text{min}_b, \text{max}_b \right)\right) \times s, 
\label{eq:quantization}
\end{gather}
where \(\mathbf V\) and $Q(\mathbf{V})$ denote the floating-point value and its dequantized value (quantization width is $b$ bit). The parameter $s = \frac{\max(\mathbf V) - \min(\mathbf V)}{2^b - 1}$, which represents the scale factor used in the quantization mapping, the interval $[{min}_b, {max}_b]$ specifies integer range.

\subsection{Problem Formulation}
\label{subsec:fomulation}

Fig.~\ref{fig:main} illustrates the overall framework of JAQ, which includes the joint search among network structure, ultra-low mixed-precision bitwidth allocation, and accelerator architecture.
The formulation of the joint optimization problem is:

\vspace{-0.2cm}

\begin{gather}
\min_{\boldsymbol{\alpha}, \boldsymbol{\beta},\boldsymbol{\gamma},\mathbf w}  
\mathcal{L}_{\text{\rm CE}} \left(\mathbf w, \text{N}(\boldsymbol{\alpha}), \text{M}(\boldsymbol{\beta}) \right) \nonumber \\
\text{s.t.} \quad \mathcal{E}_{\rm HW} \left(\text{H} \left( \boldsymbol{\gamma} \right), \text{N} \left( \boldsymbol{\alpha} \right), \text{M}(\boldsymbol{\beta}) \right) \leq C
\label{eq:mainObject}
\end{gather} 
where $\boldsymbol{\alpha}$ and $\boldsymbol{\beta}$ denote the operator architecture parameters and the bitwidth architecture parameters, respectively. 
$\boldsymbol{\gamma}$ denotes the hardware accelerator configuration. 
\(\mathbf w\) represents the weights of the NAS supernet. 
\(\text{N}(\boldsymbol{\alpha})\) indicates the network structure selected based on $\boldsymbol{\alpha}$. \(\text{M}(\boldsymbol{\beta})\) denotes the bitwidths selection for each operator according to $\boldsymbol{\beta}$. 
\(\text{H}(\boldsymbol{\gamma})\) depicts the accelerator architecture based on $\boldsymbol{\gamma}$. \(\mathcal{E}_{\rm HW}\) reflects the hardware-side performance, calculated by hardware metrics (Energy, Latency, and Area). 
% \yym{XXX}.\wmz{done}
\( \mathcal{L}_{\text{\rm CE}} \) represents the cross-entropy loss, and \(\mathcal{E}_{\rm HW}\). 
To track this optimization problem, we introduce a Lagrange multiplier \(\lambda\) for Eq~\eqref{eq:mainObject}:
\begin{equation}
\small
    \min_{\boldsymbol{\alpha}, \boldsymbol{\beta},\boldsymbol{\gamma},\mathbf w} \Big[ \mathcal{L}_{\text{\rm CE}} \left(\mathbf w, \text{N}(\boldsymbol{\alpha}), \text{M}(\boldsymbol{\beta}) \right) 
 + \lambda \mathcal{E}_{\rm HW} \left(\text{H} \left( \boldsymbol{\gamma} \right), \text{N} \left( \boldsymbol{\alpha} \right), \text{M}(\boldsymbol{\beta}) \right) \Big].  
\label{eq:mainObject} 
\end{equation}

% \vspace{-0.2cm}
\subsection{Channel-wise Sparse Quantization (CSQ)}
\label{subsec:algorithm}
\begin{figure*}[t]
    \centering

\begin{subfigure}[b]{0.3\textwidth} % 这里的 0.3\textwidth 控制 subfigure 容器的宽度
    \centering
    \includegraphics[width=0.9\linewidth, height=3.5cm]{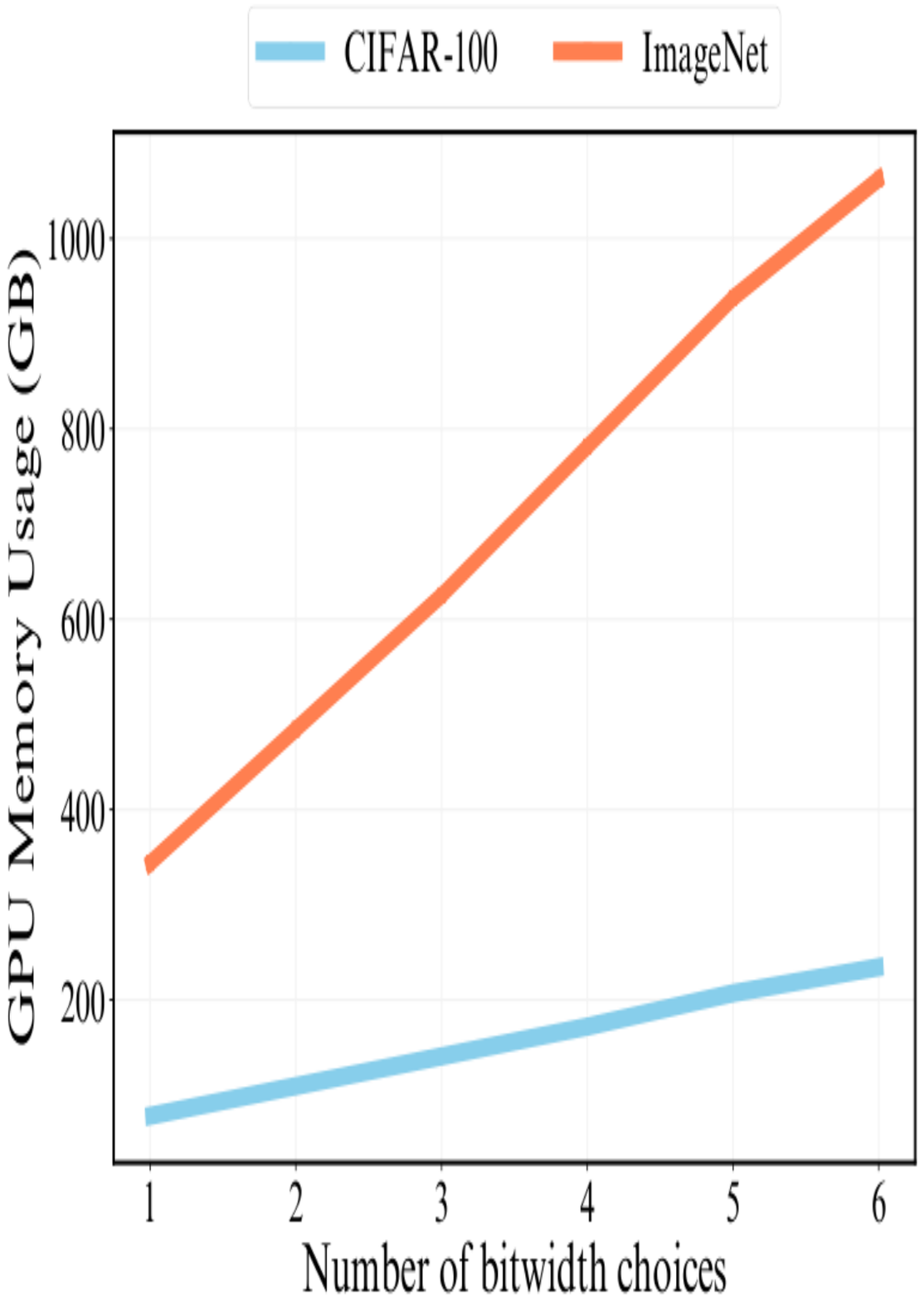} % 调整 width 和 height
    \caption{increasing bitwidth choices}
    \label{fig:gpu_use}
\end{subfigure}
    \hfill % Use hfill to push the second subfigure to the right
    \begin{subfigure}[b]{0.3\textwidth}
        \centering
        \includegraphics[width=0.9\linewidth, height=3.5cm]{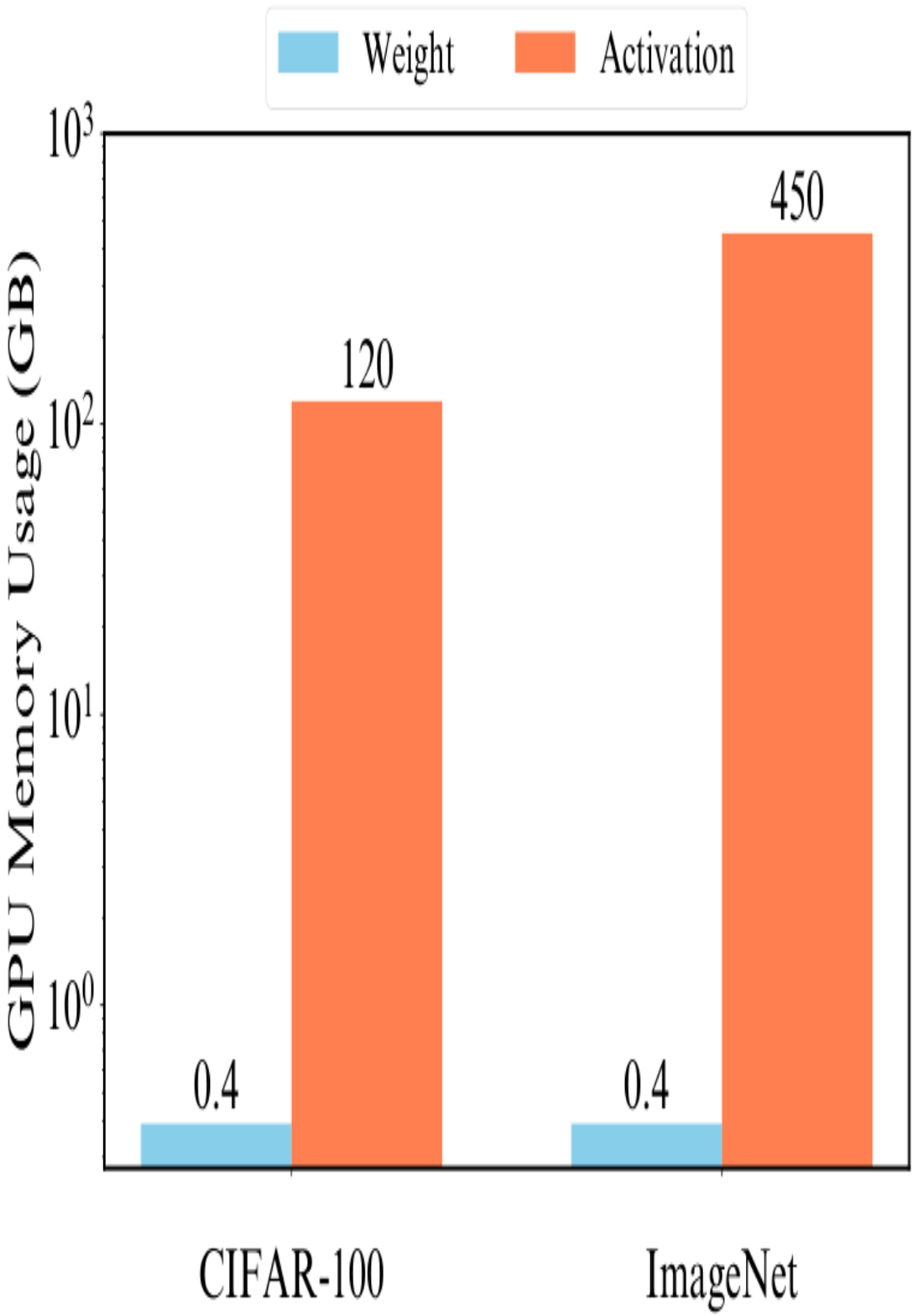} % Adjust path as needed
        \caption{comparison of w/a}
        \label{fig:act_wei}
    \end{subfigure}
    \hfill % Use hfill to push the second subfigure to the right
    % Second image
    \begin{subfigure}[b]{0.3\textwidth}
        \centering
        \includegraphics[width=0.9\linewidth, height=3.5cm]{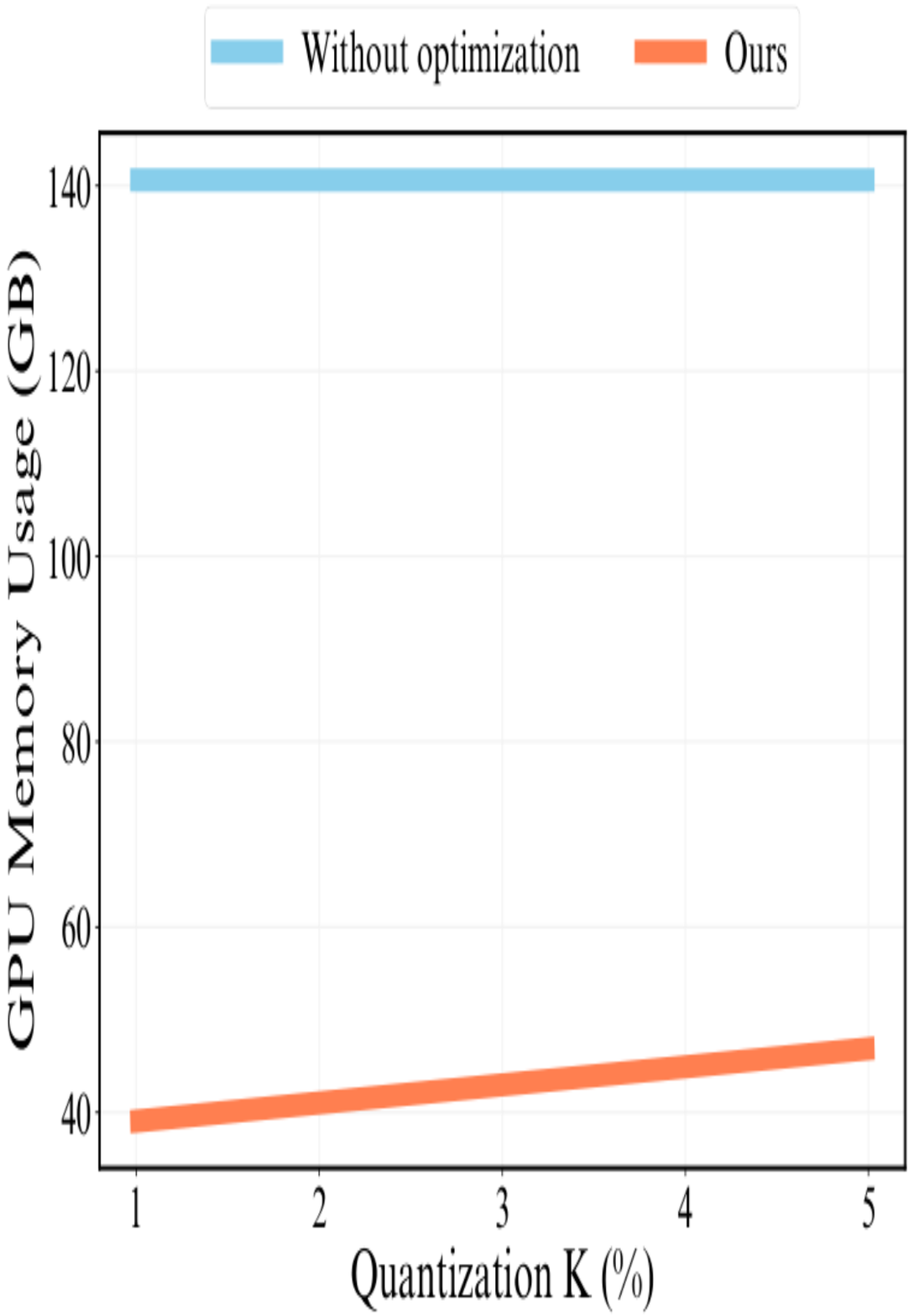} % Adjust path as needed
        \caption{memory reduction degree}
        \label{fig:gpu_compare}
    \end{subfigure}
    \caption{(a) depicts the GPU memory usage with increasing bitwidths choices on CIFAR-100 and ImageNet (batch size is 128). (b) presents the GPU memory usage during the quantization stage for weights and activations on CIFAR-100 and ImageNet (batch size 256). (c) contrasts GPU memory usage on CIFAR-100 among our work and the non-optimized baseline (batch size 256).} % General caption for the figure
    \label{fig:gpu_memory}
\end{figure*}

\paragraph{Memory Cost Bottleneck.}
DNAS segments the supernet into a series of cells. Each cell is structured as a directed acyclic graph (DAG) with several nodes (each node corresponds to a distinct operator), and each operator within the supernet must be stored in GPU memory during training.
This indicates that adding extra search dimensions in DNAS-based methods can easily lead to GPU memory overload or require to reduce the training batch size to maintain the original utilization of GPU memory.
Quantization, a memory-intensive process, involves storing numerous quantized parameters and additional quantization information. Therefore, integrating network architecture search with bit-width selection significantly amplifies the GPU memory consumption.
For example, as shown in Fig.~\ref{fig:gpu_use}, the utilization of GPU memory increases linearly with the number of available bitwidth options in each operator which is deemed unacceptable. For more detailed methods of measuring GPU memory consumption, please refer to Appendix A.

\paragraph{CSQ.}

The differentable network and bitwidths co-search framework in JAQ can be implemented by formulating as:
\begin{gather}
   {\mathbf A^{l+1}} = \sum_{i=1}^{n} \boldsymbol{\alpha}_i^{l} \cdot \tilde{\mathbf W}_i^{l} \cdot \tilde{\mathbf A}_i^{l}, \; \text{where} \quad \nonumber \\
   % \text{where} \quad 
    \tilde{\mathbf W}_i^{l} = \sum_{k=1}^{m} \boldsymbol{\beta}_{ w_{i,k}^l} \cdot Q({\mathbf W_{ik}^l}), \; \text{and} \;
    \tilde{\mathbf A}_i^{l} = \sum_{k=1}^{m} \boldsymbol{\beta}_{ a_{i,k}^l} \cdot Q({\mathbf A}^l),
    \label{eq:origin_activ}
\end{gather}

% \yym{alpha, n？}\wmz{done}
where $l$ denotes the layer index in network, and $n$ is the number of operator candidates per layer, while $m$ is the number of bit-width candidates per operator. $\tilde{W}$ and $\tilde{A}$ represent the sum of quantized weights and quantized activations under different precisions respectively. $\boldsymbol{\alpha}$ denotes the operator architecture parameters, while $\boldsymbol{\beta}_{ w_{i,k}^l}$ and $\boldsymbol{\beta}_{ a_{i,k}^l}$ are the architecture parameters of weights and activations for each precision. $Q$ represents the quantization function (Eq.~\ref{eq:quantization}).
% QQ represents the quantization function in Equation~\ref{eq:quantization}.
As illustrated in Fig.~\ref{fig:act_wei}, during the supernet training process, the memory requirement of weight quantization is trivial. 
Therefore, memory cost bottleneck in network and bitwidths co-search framework can be predominantly attributed to the quantization of activations. To alleviate this issue during supernet training, we propose channel-wise sparse quantization strategy for the quantization of activation. This can be implemented by reformulating Eq.~\ref{eq:origin_activ} as:
\begin{gather}
    \tilde{\mathbf A}_i^{l} = \left( \sum_{k=1}^{m} \boldsymbol{\beta}_{ a_{i,k}^l} \cdot Q(\mathbf A^{l}(\Omega^{l})) \right) \oplus \mathbf A^{l}(1-\Omega^{l}),
    \label{eq:Omega}
\end{gather}
where $\Omega$ is the indices of channels to be quantized and $\oplus$ denotes concatenation operation. $\mathbf A^{l}(\Omega^{l})$ represents all channels selected in $\mathbf A^{l}$ according to $\Omega^{l}$.

The core innovation of this method is to quantize only a few channels of activations during searching phase, while leaving other channels unquantized, which significantly reduces the demand on GPU memory. To achieve better search result (detailed explanation and experiments are provided in the ablation study), we need to select the most important channels from each activations. Inspired by a previous work~\cite{liu2017learning}, the scale factors in Batch Normalization (BN) can effectively represent the importance of each channel. 
\begin{gather}
\hat{z} = \frac{z_{\text{in}} - \mu_B}{\sqrt{\sigma^2_B + \epsilon}}, \quad z_{\text{out}} = \gamma \hat{z} + \beta,
\label{eq:bn}
\end{gather}
where \( z_{\text{in}} \) and \( z_{\text{out}} \) are the input and output of a BN layer, \( \mu_B \) and \( \sigma_B \) represent the mean and standard deviation of the input activations across the batch \( B \). The trainable parameters  \( \gamma \) and \( \beta \) serve as scale and shift factors respectively.

Therefore, we choose to quantize only the top \( K\% \) of the most important channels of each activations during search phase and defining:
% \yuan{fixed top k channel in for every batch?}
\begin{gather}
    \Omega^{l} \leftarrow \text{\tt TopKChannelToQuantize}(\Gamma^{l}, K) \nonumber \\
    % \quad \text{where} \quad
      \Gamma_j^{l} = \sum_{i=1}^{n} \boldsymbol{\alpha}_i^{l-1} \gamma_{ij}^{l-1} \quad j \in N^{l-1},
    % \Gamma_j^{l} = \sum_{i=1}^{n} \alpha_i^{l-1} \gamma_{ij}^{l-1} \quad j \in [1, \text{Num}_{\text{output channel}}^{l-1}],
    \label{eq:K}
\end{gather}
where $\Gamma$ is the importance indicator of each channel, and $\gamma$ represents the scale factors defined in Eq.~\ref{eq:bn}, while N is the output channels number of $l$ - 1 layer.

During the search phase, scale factors are trainable to dynamically adjust the importance indicator for each channel. Finally, as Fig.~\ref{fig:gpu_compare} demonstrates, the GPU utilization in our algorithm is significantly reduced to acceptable bounds.

\subsection{Accelerator Architecture Search}
\label{subsec:hardware}

Hardware parameters in accelerators are non-differentiable. Although it is possible to optimize these parameters using reinforcement learning~\cite{lin2021naas}, the time overhead is particularly substantial. Therefore, there is a demand for exploring efficient methods to search for these parameters.
Furthermore, compiler mapping is crucial for the latency and energy consumption of DNNs inference on accelerators. Therefore we incorporate compiler mapping optimization into the joint search framework, reducing its searching time to less than 0.15 seconds per iteration.
\begin{figure*}[h]
    \centering
    \includegraphics[width=0.95\linewidth]{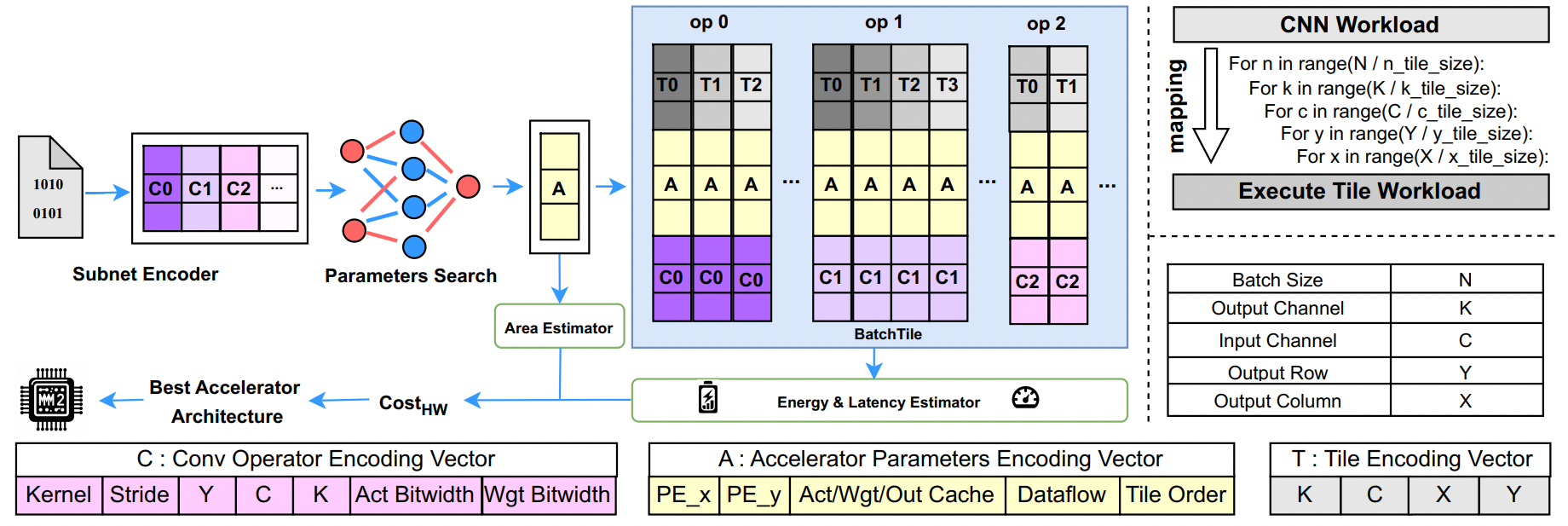}
    \caption{The overall accelerator search framework of JAQ. The right part represents the executing workload of a CNN operator after compiler mapping, which can be segmented into tiles across five dimensions. The left part displays an optimization pipeline including subnet encoder, accelerator parameters search, and the BatchTile method. The bottom section elaborates on the meanings of each field within the three distinct vectors.}
    \label{fig:hw_framework}
\end{figure*}

\vspace{-0.1cm}
\paragraph{Accelerator Search Space.}

Our accelerator search space is divided into two categories. The first category involves accelerator parameters, which include the shape and number of processing elements (PEs), the size of the on-chip cache used for storing weights, activations, and outputs, as well as the inter-connection of PEs, described as the dataflow type of the parallel dimension. The second category focuses on optimizations of the compiler mapping, including sizes of tiles and loop order of tiling.

\paragraph{Accelerator Parameters Search.}

First, we identify the current optimal subnet within the supernet and encode each operator in the subnet as the operator encoding vector in Fig.~\ref{fig:hw_framework}: Kernel, Stride, Output Row, Input Channel, Output Channel, Activation Bitwidth, and Weight Bitwidth.
Then, the encoded operators are sent into the accelerator parameters search part, constituted by five layers of residual blocks. 
The final layer maps the hidden states into seven elements in the accelerator parameters encoding Vector in Fig.~\ref{fig:hw_framework}: \( \text{PE}_x \), \( \text{PE}_y \), Activation Cache, Weight Cache, Output Cache, Dataflow Type, and Tile Order. 

Gumbel-Softmax~\cite{jang2016categorical} is used as the activation function in each classifier, ensuring that the output values closely resemble the inputs for hardware cost estimation, as well as maintaining the gradient propagation during the training stage.
\paragraph{Compiler Mapping Search: BatchTile.}
\label{subsec:compiler}

\begin{table}[t]
\vspace{-0.1cm}

\begin{center}
\resizebox{0.45\textwidth}{!}{
\renewcommand{\arraystretch}{1} % 控制换行的大小
\LARGE
\begin{tabular}{lccccccc}
% \toprule[1.2pt] 
\\  \cline{1-8} \\
\multicolumn{1}{c}{\bf \makecell{}} 
&\multicolumn{1}{c}{\bf \makecell{B/b}}
&\multicolumn{1}{c}{\bf \makecell{OW/ow}}
&\multicolumn{1}{c}{\bf \makecell{OH/oh}}  
&\multicolumn{1}{c}{\bf \makecell{IC/ic}}
&\multicolumn{1}{c}{\bf \makecell{OC/oc}}
&\multicolumn{1}{c}{\bf \makecell{Latency(ms)}}
&\multicolumn{1}{c}{\bf \makecell{Energy(mJ)}}
\\  \cline{1-8} \\
    \makecell{Best Tiling}
    &{\makecell{1/1}}
    &{\makecell{7/1}}
    &{\makecell{7/1}}
    &{\makecell{35/16}}
    &{\makecell{35/16}}
    &{\makecell{3.2}}
    &{\makecell{2.6}}
\\  \cline{1-8} \\
    \makecell{Case 0}
    &{\makecell{1/1}}
    &{\makecell{2/4}}
    &{\makecell{2/4}}
    &{\makecell{276/2}}
    &{\makecell{2/512}}
    &{\makecell{56.6}}
    &{\makecell{8.3}}
\\  \cline{1-8} \\
    \makecell{Case 1}
    &{\makecell{1/1}}
    &{\makecell{4/2}}
    &{\makecell{4/2}}
    &{\makecell{18/32}}
    &{\makecell{5/128}}
    &{\makecell{4.7}}
    &{\makecell{5.1}}
\\  \cline{1-8} \\
% \\  \bottomrule[1.2pt]
\end{tabular}
}
\vspace{-0.1cm}
\end{center}
\caption{This table presents the latency and energy for the optimal tiling method and two randomly selected tiling methods applied to a specific pair of operator and accelerator (Operator: kernel size of 5, stride of 1, output size of 7, both input and output channels at 552, with activation and weight bitwidths of 8 bits. Accelerator: PE Array dimensions of 16x16, with 384KB Act/Wgt/Out Cache sizes).}
\label{tab:tile}
\end{table}

As shown in Tab.~\ref{tab:tile}, Tile size is crucial for the model inference performance on accelerator. 
In JAQ, the accelerator is configured to process one image at a time, hence the batch size is one. 
Consequently, each operator requires tiling across four dimensions: input channel, output channel, output width, and output height.
To achieve peak performance for model inference on accelerator, optimal tile sizes for each operator should be determined during the compiler mapping stage. 
However, finding the optimal tile size for all operators in a subnet is time-consuming (around 50s), which is unfriendly to end-to-end joint search. 
To efficiently find the optimal tile size, we propose the BatchTile method. 
The BatchTile method initially encodes each operator's tiling strategies across four dimensions as illustrated in Fig.~\ref{fig:hw_framework}: Output Channel, Input Channel, Output Column, and Output Row. 
Subsequently, we concatenate accelerator parameters encoding vector, operator encoding vector of each operator, and different tile encoding vectors to form various $\langle$Operator, Accelerator Parameters, Tiling Strategy$\rangle$ pairs. 
These pairs, as different batches, are fed into the Energy \& Latency Estimator (the principle of the estimator follows~\cite{choi2021dance}) to simultaneously identify the optimal tiling strategy for each operator.
Finally, the BatchTile method reduces the entire compiler mapping search time to less than 0.15 seconds(comparison experiment is in Appendix C).

Our methodology capitalizes on hardware design principles, estimating the hardware from the perspective of area, energy, and latency metrics. Combining these three metrics, the hardware cost function included in Eq.~\ref{eq:mainObject} is:
\begin{gather}
\small
\mathcal{E}_{\text{HW}}\  = \lambda_E \cdot \text{\tt Energy} + 
\lambda_L \cdot \text{\tt Latency} + \lambda_A \cdot \text{\tt Area},
\label{eq:hw_cost}
\end{gather}
where \( \lambda_E \), \( \lambda_L \), and \( \lambda_A \) are adjustable among these cost metrics.

\paragraph{Generalizability.}
Our accelerator search methodology is general, with no prior assumptions about the types of accelerators used. As a result, it is suitable for various accelerator architectures and compiler mapping strategy. Our search methodology can be readily utilized by providing (1) a hardware cost estimator, and (2) a set of user-defined accelerator parameters and compiler mapping search space. This demonstrates the flexibility and generalizability of our search strategy.

\subsection{The Overall Joint Pipeline}
\label{subsec:joint}
JAQ consists of the search stage and the retrain stage.
The search stage integrates the channel-wise sparse quantization method into the model (network architectures and bitwidths) searching, and incorporates the BatchTile approach into the accelerator searching. 

For searching, each iteration consists of two steps. The first step is to update the weights ($\mathbf w$) in supernet, which doesn't require interfacing with the accelerator. The second step, collaborating with the accelerator, involves updating the architecture parameters ($\boldsymbol{\alpha}$ and $\boldsymbol{\beta}$) and the accelerator configuration ($\boldsymbol{\gamma}$), as defined in Eq.~\ref{eq:mainObject}. In the second step, after forward propagation in the supernet, the current optimal subnet is encoded and passed into the accelerator search framework. Then, we optimize the accelerator parameters and compiler mapping strategy. Subsequently, the \(\text{\tt Cost}_{\text{HW}}\) obtained through Eq.~\ref{eq:hw_cost} is bound to the architecture parameters, which will be updated during the backpropagation process.

For retraining, we retrain the optimal subnet obtained from the search stage. Finally, we achieve the optimal network structure and accelerator architecture, thus realizing the synergy between software and hardware design.

%% file: experiments.tex
\subsection{Experimental Settings}
\label{subsec:settings}
Our experiments are conducted on the CIFAR-10/100, and ImageNet datasets. 
In search stage, We use $80\%$ of the data to update the weights within the supernet and $20\%$ of the data for the architecture parameters. The initial learning rate is $0.01$, employing an annealing cosine learning rate schedule. The initial temperature for the Gumbel-Softmax is set to $5$. For the CIFAR-10/100 and ImageNet datasets, we search for 90 and 45 epochs on eight NVIDIA GeForce RTX 4090 GPUs, respectively. In Eq.~\ref{eq:K}, we select K as 3. In Eq.~\ref{eq:hw_cost}, $\lambda_E$, $\lambda_L$, and $\lambda_A$ are all set to 0.33.
In retrain stage, we train the subnet for 600 epochs for CIFAR-10/100 and 180 epochs for ImageNet, respectively. We employ an annealing cosine learning rate schedule, with an initial learning rate of $0.01$.

\subsection{Search Space}

We utilize FBNet~\cite{wu2019fbnet} as the network search space. 
Except for stem and head layers, it comprises 22 blocks. Each block has 9 candidate operations, including a skip choice. 
We utilize BitFusion~\cite{sharma2018bit} accelerator as the hardware template, which is a SOTA ASIC accelerator for mixed-precision models.
% For the search space of bitwidths, due to the BitFusion~\cite{sharma2018bit} accelerator supports 2, 4, 8 bits, weights and activations of each layer have three different bitwidth options.
For the search space of bitwidths, the weights and activations of each layer have three different options $\in [2,4,8]$.
For the accelerator search space, \( \text{PEx} \) and \( \text{PEy} \) are selectable within a range of 3 to 64. The cache sizes for weights, activations, and outputs are configurable in increments of 16KB, ranging from 64KB to 528KB, offering 30 distinct choices. We choose three types of dataflows: Weight Stationary (WS)~\cite{jouppi2017datacenter}, Output Stationary (OS)~\cite{du2015shidiannao}, and Row Stationary (RS)~\cite{chen2016eyeriss}. For each operator, there are 120 possible permutations of the tile order across five dimensions: batch size, input channel, output channel, output height, and output width. For tile sizes, we set the batch size to only one, while in other dimensions, the tile size can vary from \(2^0\) to \(2^n\) (The maximum value of n is 10).

\subsection{Co-exploration Results}

\begin{table}[t]

\renewcommand{\arraystretch}{1} % 控制换行的大小
\small
\begin{subtable}{\textwidth}
\resizebox{0.47\textwidth}{!}{
\begin{tabular}{lcccccccc}
\hline
\multicolumn{1}{c}{\bf \makecell{}} 
&\multicolumn{2}{c|}{\bf \makecell{$\lambda$ = 0.004}}
&\multicolumn{2}{c|}{\bf \makecell{$\lambda$ = 0.002}}  
&\multicolumn{2}{c|}{\bf \makecell{$\lambda$ = 0.001}}
&\multicolumn{2}{c}{\bf \makecell{$\lambda$ = 0.0005}}
\\ \cline{1-9}
\multicolumn{1}{c}{\bf \makecell{}} 
&\multicolumn{1}{c}{\bf \makecell{ACC}}
&\multicolumn{1}{c|}{\bf \makecell{EDAP}}
&\multicolumn{1}{c}{\bf \makecell{ACC}}  
&\multicolumn{1}{c|}{\bf \makecell{EDAP}}
&\multicolumn{1}{c}{\bf \makecell{ACC}}
&\multicolumn{1}{c|}{\bf \makecell{EDAP}}
&\multicolumn{1}{c}{\bf \makecell{ACC}}
&\multicolumn{1}{c}{\bf \makecell{EDAP}}
\\ \hline
    \makecell{Auto-nba}
    &{\makecell{82.847}}
    &{\makecell{10}}
    &{\makecell{89.643}}
    &{\makecell{12.8}}
    &{\makecell{86.597}}
    &{\makecell{20}}
    &{\makecell{86.677}}
    &{\makecell{26}}
\\ \hline
    \makecell{\textbf{Ours}}
    &{\makecell{91.081}}
    &{\makecell{11.8}}
    &{\makecell{92.163}}
    &{\makecell{12.4}}
    &{\makecell{91.895}}
    &{\makecell{17.6}}
    &{\makecell{92.183}}
    &{\makecell{30}}
\\ \hline
\end{tabular}
}
\end{subtable}

\hspace{3.5cm} % 右移2cm
\textbf{\fontsize{8}{10}\selectfont (a) CIFAR10} % 调整字体为粗体，字号为10pt

\vspace{0.2cm}
\begin{subtable}{\textwidth}
\resizebox{0.47\textwidth}{!}{
\begin{tabular}{lcccccccc}
\hline
\multicolumn{1}{c}{\bf \makecell{}} 
&\multicolumn{2}{c|}{\bf \makecell{$\lambda$ = 0.004}}
&\multicolumn{2}{c|}{\bf \makecell{$\lambda$ = 0.002}}  
&\multicolumn{2}{c|}{\bf \makecell{$\lambda$ = 0.001}}
&\multicolumn{2}{c}{\bf \makecell{$\lambda$ = 0.0005}}
\\ \cline{1-9}
\multicolumn{1}{c}{\bf \makecell{}} 
&\multicolumn{1}{c}{\bf \makecell{ACC}}
&\multicolumn{1}{c|}{\bf \makecell{EDAP}}
&\multicolumn{1}{c}{\bf \makecell{ACC}}  
&\multicolumn{1}{c|}{\bf \makecell{EDAP}}
&\multicolumn{1}{c}{\bf \makecell{ACC}}
&\multicolumn{1}{c|}{\bf \makecell{EDAP}}
&\multicolumn{1}{c}{\bf \makecell{ACC}}
&\multicolumn{1}{c}{\bf \makecell{EDAP}}
\\ \hline
    \makecell{Auto-nba}
    &{\makecell{60.169}}
    &{\makecell{4}}
    &{\makecell{52.837}}
    &{\makecell{6.2}}
    &{\makecell{56.468}}
    &{\makecell{14}}
    &{\makecell{48.542}}
    &{\makecell{18}}
\\ \hline
    \makecell{\textbf{Ours}}
    &{\makecell{72.440}}
    &{\makecell{2.6}}
    &{\makecell{72.956}}
    &{\makecell{7.8}}
    &{\makecell{73.264}}
    &{\makecell{13.4}}
    &{\makecell{73.651}}
    &{\makecell{14.2}}
\\ \hline
\end{tabular}
}
\end{subtable}

\hspace{3.5cm} % 右移2cm
\textbf{\fontsize{8}{10}\selectfont (b) CIFAR100} % 调整字体为粗体，字号为10pt

\vspace{0.2cm}
\begin{subtable}{\textwidth}
\resizebox{0.47\textwidth}{!}{
\begin{tabular}{lcccccccc}
\hline
\multicolumn{1}{c}{\bf \makecell{}} 
&\multicolumn{2}{c|}{\bf \makecell{$\lambda$ = 0.005}}
&\multicolumn{2}{c|}{\bf \makecell{$\lambda$ = 0.002}}  
&\multicolumn{2}{c}{\bf \makecell{$\lambda$ = 0.001}}
\\ \cline{1-7}
\multicolumn{1}{c}{\bf \makecell{}} 
&\multicolumn{1}{c}{\bf \makecell{ACC}}
&\multicolumn{1}{c|}{\bf \makecell{EDAP}}
&\multicolumn{1}{c}{\bf \makecell{ACC}}  
&\multicolumn{1}{c|}{\bf \makecell{EDAP}}
&\multicolumn{1}{c}{\bf \makecell{ACC}}
&\multicolumn{1}{c}{\bf \makecell{EDAP}}
\\ \hline
    \makecell{Auto-nba}
    &{\makecell{62.423}}
    &{\makecell{32.48}}
    &{\makecell{61.781}}
    &{\makecell{253}}
    &{\makecell{62.787}}
    &{\makecell{503.1}}
\\ \hline
    \makecell{\textbf{Ours}}
    &{\makecell{69.132}}
    &{\makecell{26.238}}
    &{\makecell{69.473}}
    &{\makecell{230.1}}
    &{\makecell{70.197}}
    &{\makecell{498.6}}
\\ \hline
\end{tabular}
}
\end{subtable}

\hspace{3.5cm} % 右移2cm
\textbf{\fontsize{8}{10}\selectfont (c) ImageNet} % 调整字体为粗体，字号为10pt

\caption{Comparisons between our method and the baseline(Auto-nba~\cite{fu2021auto}) on three distinct datasets: CIFAR-10, CIFAR-100, and ImageNet. EDAP ($J \cdot s \cdot m^{2} \cdot 10^{-18}$) stands for the Energy-Delay-Area Product, which is a common hardware metric.}
\label{tab:main_experiment}
\end{table}

Compared with previous joint search framework~\cite{fu2021auto}, we conduct experiments on the CIFAR-10, CIFAR-100, and ImageNet (ILSVRC2012) datasets. In various comparative experiments, we adjusted $\lambda$ parameter in Eq.~\ref{eq:mainObject} to achieve different balances between accuracy and hardware cost. Specifically, on the CIFAR-10 and CIFAR-100 datasets, the value of $\lambda$ is set to 0.0005, 0.001, 0.002, and 0.004, while on ImageNet, it is set to 0.001, 0.002, and 0.005. As shown in Tab.~\ref{tab:main_experiment}, the experiments reveal that our method significantly outperforms baseline in low-bit joint search tasks.

\begin{table}[t]
\centering

\label{search time}
\resizebox{0.45\textwidth}{!}{ % 控制表格的大小
\renewcommand{\arraystretch}{0.7} % 控制行距
\LARGE
\begin{tabular}{ccccc}
\\  \cline{1-5} \\
% \toprule[1.2pt]
\multicolumn{1}{c}{\bf \makecell{Method}} 
&\multicolumn{1}{c}{\bf \makecell{Network}}
&\multicolumn{1}{c}{\bf \makecell{Bitwidth}}  
&\multicolumn{1}{c}{\bf \makecell{Accelerator}}
&\multicolumn{1}{c}{\bf \makecell{Search Time}}
\\  \cline{1-5} \\
\makecell{NAAS~\cite{lin2021naas}}
&{\makecell{---}}
&{\makecell{---}}
&{\makecell{\ding{51}}}
&{\makecell{1200}}
\\ \cline{1-5} \\
\makecell{OQAT~\cite{shen2021once}}
&\makecell{\ding{51}}
&\makecell{\ding{51}}
&\makecell{---}
&\makecell{1200}
\\ \cline{1-5} \\
\makecell{BatchQuant~\cite{bai2021batchquant}}
&\makecell{\ding{51}}
&\makecell{\ding{51}}
&\makecell{---}
&\makecell{1800}
\\ \cline{1-5} \\
\makecell{Auto-nba~\cite{fu2021auto}}
&\makecell{\ding{51}}
&\makecell{\ding{51}}
&\makecell{\ding{51}}
&\makecell{180}
\\ \cline{1-5} \\
\makecell{\textbf{Ours}}
&\makecell{\ding{51}}
&\makecell{\ding{51}}
&\makecell{\ding{51}}
&\makecell{160}
\\ \cline{1-5} \\
% \\ \bottomrule[1.2pt]
\end{tabular}
}
\vspace{0.1cm}
\caption{Comparison of search space and search time(GPU hours) between JAQ and other works on the ImageNet dataset.}
\label{tab:search time}
\end{table}
% \vspace{-0.3cm}
To demonstrate the efficiency of the JAQ, we conduct comparative analyses with other search frameworks. 
As shown in Tab.~\ref{tab:search time}, NAAS~\cite{lin2021naas} employs reinforcement learning (RL) to jointly search network structures and accelerator architectures. 
OQAT~\cite{shen2021once} and BatchQuant~\cite{bai2021batchquant} utilize a one-shot approach for joint searching of network structures and bitwidths. 
In contrast, Auto-nba~\cite{fu2021auto} and our work both present a triple search framework, but our work achieves better search efficiency within a large search space.

\begin{table}[t]
\centering
\label{tab:sensitivity_analysis}
\resizebox{0.47\textwidth}{!}{ % 控制表格的大小
\renewcommand{\arraystretch}{0.4} % 控制行距
\begin{tabular}{cccccccc}
% \toprule[1.2pt]
\\ \cline{1-8} \\
\multicolumn{1}{c}{ \makecell{}} 
&\multicolumn{1}{c}{ \makecell{$\lambda_E$}}
&\multicolumn{1}{c}{ \makecell{$\lambda_L$}}
&\multicolumn{1}{c}{ \makecell{$\lambda_A$}}  
&\multicolumn{1}{c}{ \makecell{Acc}}
&\multicolumn{1}{c}{ \makecell{Latency (ms)}}
&\multicolumn{1}{c}{ \makecell{Energy (mJ)}}
&\multicolumn{1}{c}{ \makecell{Area (mm$^2$)}}
\\ \cline{1-8} \\
\makecell{Latency-Sensitive}
&\makecell{0.1}
&\makecell{0.8}
&\makecell{0.1}
&\makecell{75.099}
&\makecell{1.94}
&\makecell{1.52}
&\makecell{1.36}
\\ \cline{1-8} \\
\makecell{Area-Sensitive}
&\makecell{0.1}
&\makecell{0.1}
&\makecell{0.8}
&\makecell{73.641}
&\makecell{2.74}
&\makecell{2.43}
&\makecell{0.69}
\\ \cline{1-8} \\
% \\ \bottomrule[1.2pt]
\end{tabular}
}
\vspace{0.1cm}
\caption{Different hardware sensitivity experiments on CIFAR-100 dataset.}
\label{tab:hardware sensitivity}
\end{table}

Hardware design must take into account the actual requirements for energy, latency, and area. Some accelerators are specifically designed to minimize power consumption and latency for deployment on embedded platforms, while others are produced to occupy a tiny area for integration into System on Chips (SoCs). The JAQ method can satisfy the sensitivity of a specific metric by adjusting the parameters in Eq.~\ref{eq:hw_cost}. 
As shown in Tab.~\ref{tab:hardware sensitivity}, for instance, increasing the $\lambda_L$ results in a low latency in the final result. Conversely, increasing the $\lambda_A$ leads to a tiny area for the accelerator. Overall, this indicates that by adjusting the cost hyperparameters, JAQ can achieve a desired solution.

\vspace{0.1cm}
\subsection{Ablation Studies}

Under low bit search condition, to demonstrate the effectiveness of the channel-wise sparse quantization algorithm in addressing GPU memory bottleneck problem, we contrast JAQ with a previous work (Auto-nba~\cite{fu2021auto}) tackling the same problem. Auto-nba introduces a method called heterogeneous sampling which employs the Straight-Through Estimator (STE)\cite{bengio2013estimating} to mask the quantization operation during updating weight parameters. While updating architecture parameters, it employs hard Gumbel-Softmax to active only one bitwidth choice to save GPU memory. 
However, this method encounters two severe drawbacks under low bit search condition. First, as shown in Appendix B Fig.~\ref{fig:parameter coupling}, the architecture parameters of the operators suffer from significant parameter coupling during training, making it challenging to distinguish them effectively. 
Second, without any constraint, each bitwidth allocation will most likely converge to the maximum value within the candidate range, rather than selecting low bitwidths that severely impact performance.
Yet, as depicted in Appendix B Fig.~\ref{fig:misguided search}, many operators ultimately select the 2-bit configuration, leading to a serious misguided search.

\begin{table}[t]
\centering
\resizebox{0.45\textwidth}{!}{ % 控制表格的大小
\renewcommand{\arraystretch}{0.4} % 控制行距
\begin{tabular}{ccc}
\\ \cline{1-3} \\
% \toprule[1.2pt]
\multicolumn{1}{c}{\bf \makecell{Method}} 
&\multicolumn{1}{c}{\bf \makecell{Misguided Search (\%)}}
&\multicolumn{1}{c}{\bf \makecell{Top-1 Accuracy }}
\\ \cline{1-3} \\
\makecell{Auto-nba~\cite{fu2021auto}}
&\makecell{40}
&\makecell{55.704}
\\ \cline{1-3} \\
\makecell{Channel 0}
&\makecell{5}
&\makecell{64.355}
\\ \cline{1-3} \\
\makecell{\textbf{Ours(K=1)}}
&\makecell{0}
&\makecell{65.377}
\\ \cline{1-3} \\
\makecell{\textbf{Ours(K=5)}}
&\makecell{0}
&\makecell{65.863}
\\ \cline{1-3} \\
% \\ \bottomrule[1.2pt]
\end{tabular}
}
\vspace{0.1cm}
\caption{Comparison of different methods for the joint search of network structures and the allocation of 2, 3, and 4-bit bitwidths without any constraint on the CIFAR-100 dataset.}
\label{tab:hhhhh}
\end{table}

Furthermore, we conduct joint search of network structures and bitwidths allocation without any constraint. As shown in Tab.~\ref{tab:hhhhh}, in the first experiment, Auto-nba utilizes heterogeneous sampling to address memory explosion but suffers from severe misguided search, only achieving 55.7 top-1 accuracy. In the second experiment, we also employ the channel-wise concept but quantize only the first channel of each activation during the search process. This approach still suffers from 5\% misguided searches, indicating that fixing the selection of channels is inappropriate. Instead, selecting important channels within each layer is preferable. The third and fourth experiments implement our channel-wise sparse quantization algorithm, setting $K$ in Eq.~\ref{eq:K} to 1 and 5, respectively. Using Eq.~\ref{eq:Omega}, we selectively quantize the most important channels, effectively eliminating misguided search problem and achieving significantly higher accuracy than the previous work.

\subsection{Visualization}
% \vspace{-0.2cm}
\begin{figure}[t]
    \centering
    \includegraphics[width=3.2in, height=1.5in]{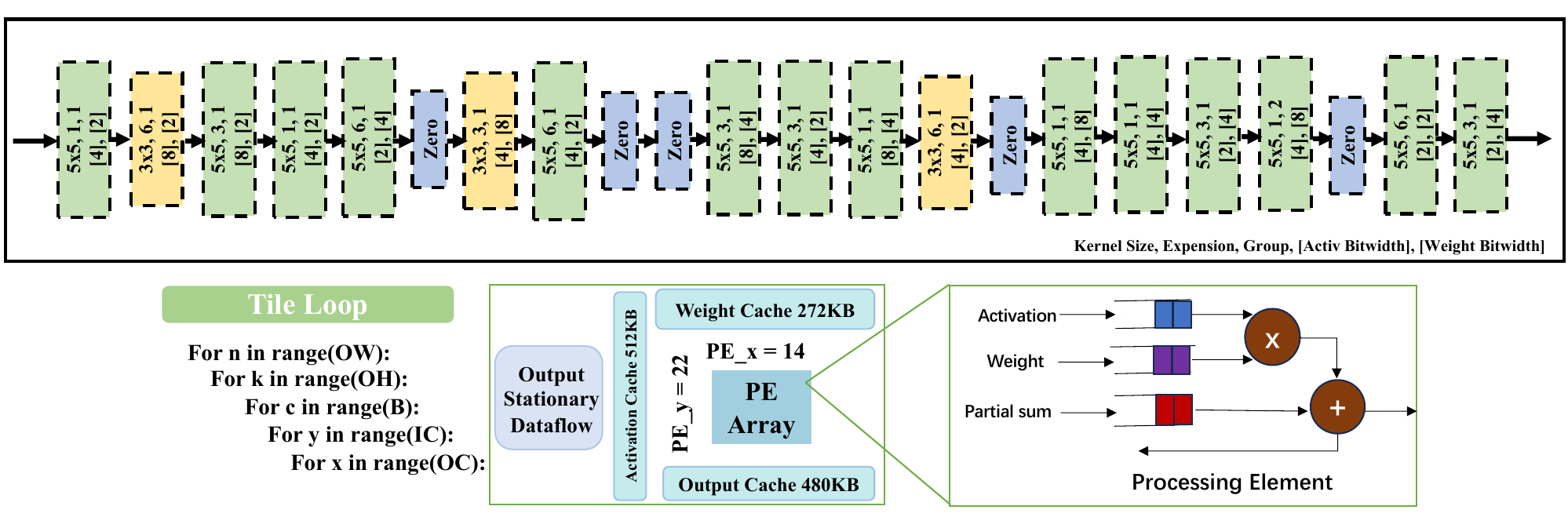}
    \caption{Visualization of searched network, bitwidths and accelerator on CIFAR-100.}
    \label{fig:visual}
\end{figure}
% \vspace{-0.3cm}
Fig.~\ref{fig:visual} indicates that convolutions with the kernel size of 5 are more compatible with the JAQ accelerator architecture, and larger kernel sizes can achieve higher accuracy with low bitwidth. Because there are more activations and outputs than weights, they are allocated a larger cache size in the search result. The output stationary dataflow is particularly well-suited to the network structure of JAQ, providing superior hardware performance.

%% file: conclusion.tex
In this paper, we present JAQ, which is the first to implement joint optimization across three dimensions: network structure, ultra-low mixed-precision bitwidths, and accelerator architecture. By addressing the challenges of memory explosion and search overhead of accelerator architecture, JAQ enables efficient joint optimization within a vast search space. When benchmarking with SOTA works, we achieve superior performance. We believe that JAQ can provide inspiration and support to the field of software-hardware co-design.